\documentclass[12pt]{spieman}  % 12pt font required by SPIE;
\usepackage{amsmath,amsfonts,amssymb}
\usepackage{graphicx}
\usepackage{setspace}
\usepackage{multirow}
\usepackage{tocloft}
\usepackage{soul}
\title{Str-L Pose: Integrating Point and Structured Line for Relative Pose Estimation in Dual-Graph}

\author[a,b]{Zherong Zhang}
\author[a,b,*]{Chunyu Lin}
\author[a,b]{Shujuan Huang}
\author[a,b]{Shangrong Yang}
\author[a,b]{Yao Zhao}
\affil[a]{With Institute of Information Science, Beijing Jiaotong University, Beijing 100044, China}
\affil[b]{Beijing Key Laboratory of Advanced Information Science and Network Technology, Beijing 10044, China}

\cftpagenumbersoff{figure}
\cftpagenumbersoff{table} 
\begin{document} 
\maketitle
\begin{abstract}
Relative pose estimation is crucial for various computer vision applications, including Robotic and Autonomous Driving. Current methods primarily depend on selecting and matching feature points prone to incorrect matches, leading to poor performance. Consequently, relying solely on point-matching relationships for pose estimation is a huge challenge. 
To overcome these limitations, we propose a Geometric Correspondence Graph neural network that integrates point features with extra structured line segments. This integration of matched points and line segments further exploits the geometry constraints and enhances model performance across different environments. We employ the Dual-Graph module and Feature Weighted Fusion Module to aggregate geometric and visual features effectively, facilitating complex scene understanding. We demonstrate our approach through extensive experiments on the DeMoN and KITTI Odometry datasets. The results show that our method is competitive with state-of-the-art techniques.
\end{abstract}

% Include a list of up to six keywords after the abstract

% Include email contact information for corresponding author
{\noindent \footnotesize\textbf{*}Chunyu Lin,  \linkable{cylin@bjtu.edu.cn} }

\begin{spacing}{2}   % use double spacing for rest of manuscript

\section{Introduction}
In computer vision, accurately estimating the relative poses between two frames is fundamental for various applications \cite{r62}, such as Structure-from-Motion (SfM) and Simultaneous Localization and Mapping (SLAM). These essential tasks involve determining the six-degree-of-freedom (6-DoF) transformation, capturing both the rotation and translation of a target frame relative to a reference frame within the complexity introduced by camera movement.

The methodologies addressing this challenge are typically categorized into two main approaches: direct methods and feature-based matching methods. Direct methods, such as \cite{r1,r2,r3,r4}, aim to find the optimal solution by minimizing photometric discrepancies based on the assumption of luminance consistency across frames. Feature-based matching methods \cite{r5,r6} infer camera motions by identifying and matching salient feature points across image sequences. Although these methodologies are fundamental to the field, they have notable limitations. Specifically, they are susceptible to feature mismatches \cite{r5} and often require additional mechanisms, such as Random Sample Consensus (RANSAC) for outlier rejection and Bundle Adjustment (BA) \cite{r7} to mitigate their vulnerabilities.

The emergence of deep learning has marked the beginning of a new phase in pose estimation, with models \cite{r8,r9,r10,r11,r12,r13,r14,r15,r16,r17,r63} demonstrating significant advancements. However, a critical bottleneck persists in environments characterized by low texture or repetitive patterns, where the scarcity or mismatches of 2D feature points can severely undermine the accuracy of the estimation \cite{r6,r14,r18,r19}. This issue demonstrates the limitations of relying solely on feature points and underscores the necessity of incorporating more holistic and structured scene representations into pose estimation frameworks. 

To address the limitations of 2D matching feature points, various methods such as \cite{r20,r21,r22} use depth information as supplementary scene features or replace 2D feature points with depth data. However, The use of depth presents two main challenges: first, the accuracy of depth data can be compromised by the limited resolution of current deep learning methods or depth sensors, leading to errors; second, depth information alone may lack the structural details needed for precise pose estimation in complex scenes. Additionally, relying heavily on 3D data increases computational complexity and may prove unfeasible in environments with limited resources.

Therefore, our method proposes using line features from the scene as supplement point features, providing a more comprehensive representation of the structural scene. Line segment information is essential as it provides robust structural cues that can complement the often sparse and unreliable point features in challenging environments. By integrating line features, we can address specific problems, such as improving pose estimation accuracy in low-texture or repetitive pattern scenes where point features alone are insufficient. Moreover, leveraging 2D data instead of 3D data reduces computation and complexity.

Our methodology focuses on an integrated \underline{Str}uctured \underline{L}ine Segment \underline{Pose} Estimation framework (Str-L Pose), which incorporates line and point features for the pose estimation. Inspired by recent advancements in graph neural networks (GNNs) \cite{r23}, known for their effectiveness in feature inference and matching, our approach combines point and line features with the geometric and topological structures of the graph. Unlike traditional methods that rely heavily on 3D data, this architecture is designed to extract and synthesize scene structure cues from 2D planes, enabling the integration of structured line and point features.

Visual information plays a critical role in understanding and interpreting a scene. It helps capture the intricate details of objects and their spatial relationships, enhancing the precision of pose estimation. To effectively integrate geometric features with visual features, we propose a Dual-Graph Neural Network framework, which includes a Geometric Correspondence Graph and a Geometry-Guided Visual Graph.
 
The Geometric Correspondence Graph integrates line and point features by using matching feature points and matching line segments, with line segments represented by two matching endpoints. The Geometry-Guided Visual Graph further improves the accuracy of pose estimation by obtaining fine-grained spatial details through coordinate sampling of image information. Our network architecture processes matching point coordinates, line coordinates, and pairs of images as inputs, translating all coordinates into high-dimensional graph node features. In the Geometric Correspondence Graph, nodes are categorized into two groups: ordinary matching points (P) and endpoints of matching line segments (L). The structural properties of line segments facilitate the interaction between points and enhance the overall estimation process.

To optimize the integration of geometric and visual features, we introduce a Feature Weighted Fusion Module. This module assigns weights to each node in the Dual-Graph structure, ensuring that the most relevant information is emphasized during the pose estimation process. By dynamically adjusting weights, this module reduces the impact of noise and inaccurate feature matching, improving the reliability and accuracy of pose estimation.

In summary, our contributions are as follows:
\begin{itemize}
    \item We propose the Str-L Pose, a novel network architecture that leverages the structured features of line segments with point features, using the geometric and topological principles inherent in Graph Neural Networks (GNN) for relative pose estimation. This integration provides a more comprehensive scene representation, significantly improving pose estimation accuracy across various environments.
    \item We introduce a unique Dual-Graph architecture that utilizes geometric features to guide the learning of visual features, enriching the model's understanding of the scene. This architecture highlights our approach to synthesizing geometric and visual data for a holistic understanding of relative poses.
    \item Our Feature Weighted Fusion Module combines geometric and visual features, assigning variable weights to optimize the integration process. This module plays a critical role in reducing the impact of noise and inaccuracies in feature matching, enhancing the reliability and precision of the pose estimation.
\end{itemize}

\section{Related work}

The exploration of relative pose estimation using deep learning techniques has advanced significantly in recent years, driven by the foundational achievements of PoseNet \cite{r8}, which employed convolutional neural networks for this task. The evolution of this domain has witnessed a range of approaches, both supervised \cite{r10,r11,r12,r13,r14,r15,r16,r17} and unsupervised learning \cite{r22,r25,r26,r27} methods, each contributing to the progress of this field.
\subsection{Learning in Pose Estimation}
The geometric methods for calculating relative poses through image-matching algorithms have led to numerous innovations in the field of pose estimation within deep learning. These methods \cite{r11,r12,r13} leverage geometric principles and have been incorporated skillfully into network designs. DeepSFM + RUMs \cite{r66} as a follow-up to DeepSFM \cite{r11}, a residual motion depth update module is added to handle dynamic scenes based on the original method of iterative refinement of depth and pose. The adoption of pose graphs, inspired by SLAM's graph optimization back-end, has propelled the development of Graph Neural Network (GNN) methodologies. Xue et.al. \cite{r28} introduced a GNN method for camera relocalization using multi-view images. Following this, Zhuang et al. \cite{r14} employed GNN to merge network solutions with geometric solutions for uncertainty fusion, achieving enhanced performance. PoserNet \cite{r29} abstracted object bounding boxes in GNN to sparsely connect multiple views, refining camera relative pose estimation. SEM-GAT \cite{r30} guided the aggregation learning of GNN by recognizing semantic relationships between nodes. 

Additionally, PoseDiffusion \cite{r57} models the conditional distribution of camera pose for a given input image using traditional polar geometric constraints and diffusion models. Chen et al. \cite {r58} used a novel parameterized representation to estimate the discrete distribution in the 5D relative pose space, aiming to improve the camera pose regression. Cho et al. \cite {r59} propose a depth photo-geometric loss to solve the scene generalization problem in camera pose estimation. SitPose \cite{r60} uses a twin convolutional transformer model SiTPose for direct regression of relative camera pose. ContextAVO \cite {r61} uses a context guidance and refinement module to recover camera motion. Fu et.al. \cite {r65} proposed a strategy for the elimination of short-term coarse to fine dynamic elements based on the checking of the state of motion (MSC) and the updating of the feature points (FPU) to better adapt to dynamic scenes. These methods aim to improve geometric constraints, parameterized representation, loss functions, and network models to enhance the accuracy and generalization ability of camera pose estimation.

% Fu et al. (2019) proposed a strategy for coarse-to-fine short-term dynamic element elimination based on motion state checking (MSC) and feature point updating (FPU) to better adapt to dynamic scenes.

Our approach, which integrates structured line and point features in an attention-based dual graph framework, is rooted in these advances. It considers geometric understanding and graph-based modeling to enhance pose estimation precision in complex scenarios.

Recently, tasks estimating relative pose have often been integrated with other related downstream tasks for joint learning, such as using depth prediction to impose geometric constraints on pose estimation. 

Zhou et al. \cite{r22} initially proposed a classic unsupervised framework for monocular depth and pose estimation. Subsequently, DualRefine \cite{r20} iteratively refined depth and pose estimation through epipolar sampling, leveraging accurate depth estimation and features for precise pose updates. R3d3 \cite{r33} refines the iterative process between spatio-temporal geometric estimation across multiple frames and monocular depth. GeoNet \cite{r21} and Un-VDNet \cite{r64} developed an unsupervised learning framework to jointly learn monocular depth, optical flow, and camera pose. Other downstream tasks include DytanVO \cite{r34}, which perfected the task of camera self-motion and motion segmentation in a single framework. OfVL-MS \cite{r32} predicted camera poses across scenes by anticipating scene coordinates with inherent uncertainties. Moran et al. \cite{r31} concurrently recovered camera poses and 3D scene structures using a set of point trajectories. XVO \cite{r67} uses segmentation, streaming, depth, and audio-assisted prediction tasks to promote the generalized representation of visual odometry. SCIPaD \cite{r68} employs a position hint aggregator to merge 3D spatial information from depth prediction with 2D feature flows and enhances pose estimation accuracy by integrating a confidence-aware feature flow estimator and a hierarchical position embedding injector.

It is worth noting that while these studies advance relative pose estimation by integrating geometric and scale information, they often overlook the potential of exclusively leveraging 2D scene features and image priors. In contrast, our approach aims to fill this gap by prioritizing 2D data.

\subsection{Line Localization}
In visually challenging environments, supplementing point features with line features has proven to be a robust strategy for enhancing visual localization and pose estimation accuracy. This combination, particularly evident in the SLAM domain, underscores the utility of diverse feature types to address limitations in environments sparse in distinctive features. In SLAM, PL-SLAM \cite{r35} utilized both point and line features throughout the SLAM process, defining line reprojection errors to address performance declines in environments with scarce point features. SuperLine \cite{r37}, Gao et al. \cite{r38} learned line segment detection, matching, and visual SLAM tasks to obtain more robust line segment features. Li et.al. \cite{r36} leveraged the structural regularities of the natural world for line-based SLAM.

Our research uses the topological structure of the graph and the matching line and point features and pioneers the use of a dual-graph structure to combine geometric features with visual features, which is different from the above methods.
\begin{figure*}[htbp]
\centering
\includegraphics[width=\textwidth]{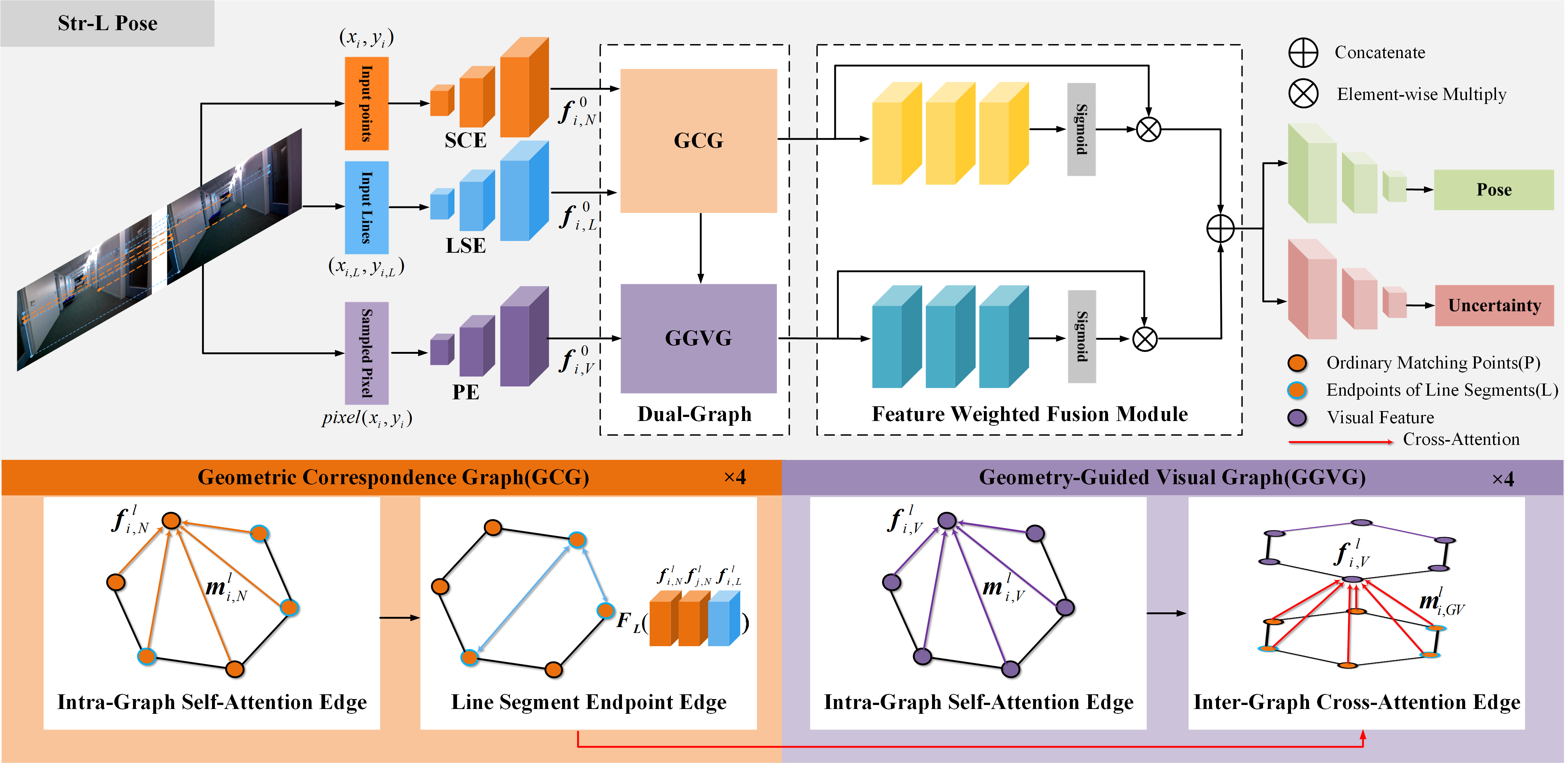}
\caption{\label{Fig1}The architecture of our proposed Str-L Pose. It contains three components: Feature Encoder(SCE, LSE, PE), Dual-Graph Architecture, and Feature Weighted Fusion Module. In the Dual-Graph architecture, the Geometric Correspondence Graph employs spatial coordinate encoding to articulate geometric relationships and facilitate accurate pose estimation through structured line segment integration, while the Geometry-Guided Visual Graph extracts and processes visual features from the input image. Feature Weighted Fusion Module harmonizes geometric and visual features.}
\label{fig:twocolumn}
\end{figure*}
\section{Method}

As illustrated in Fig.\ref{Fig1}, Str-L Pose employs a pair of images $A$ (reference image) and $B$ (target image), along with coordinates of matching points and matching line segments from both images, as inputs. Our objective is to determine the relative pose of the image pair $A$ and $B$, which includes rotation $R\in SO(3)$ and translation $t\in \mathbb{R}^3$. 

The coordinates of matching points are denoted as $(x_r, y_r)$ for the reference image and $(x_t, y_t)$ for the target image. Similarly, the matching segments are defined by two endpoints: $(x_{r,L}, y_{r,L})$, $(x_{r,L}^{'}, {y_{r,L}^{'}})$ for the reference image and $(x_{t,L}, y_{t,L})$, $(x_{t,L}^{'}, {y_{t,L}^{'}})$ for the target image.

To acquire the matching points and line segments, we use the state-of-the-art GlueStick \cite{r23} network. Our method then leverages an attention-based Geometric Correspondence Graph (GCG) Neural Network to correlate multiple matching points and line segments, generating geometric structure features. 

In parallel, we construct a Geometry-Guided Visual Graph (GGVG). From images $A$ and $B$, we sample features based on all matching coordinates. These sampled features are fed into an attention-based GGVG to extract visual features of the images.

Finally, the Feature-Weighted Fusion Module combines these geometric and visual features to generate the relative poses of the two frames.

\subsection{Basic Graph Neural Networks}
A basic graph structure G is represented as $G=<N, E>$, consisting of nodes $N$ and edges $E$. Each node $N_i$ has an initial encoding feature $f_{i,N}^{0}$. If two nodes are connected, they share an edge, which may also have initial edge encoding features $f_{i,E}^{0}$. Our method uses edges solely for message passing between connected nodes without considering their encoding features.

Graph node features are updated layer by layer through message passing and state-updating functions. At the $l$-th layer, if nodes $i$ and $j$ are connected, their message passing described by:
\begin{equation}
m^l_{ij} = M_{ij}(f^{l}_{i},f^{l}_{j}).\label{eq1}
\end{equation}
Here, $f^{l}_{i}$ and $f^{l}_{j}$ represent the feature of nodes $i$ and $j$, $M_{ij}$ denote the message passing operations.

The update of the state of node $i$ is the aggregation of messages from all its adjacent nodes:

\begin{equation}
f^{l+1}_{i} = U_i(f^{l}_{i},m^{l}_{ij}).\label{eq2}
\end{equation}
Here, $f^{l+1}_{i}$ represents the new node feature after message updates, $U_i$ denotes the state update operations. Eq. \ref{eq1} and Eq. \ref{eq2} show how node $i$ updates its state after receiving a message from connected node $j$.

\subsection{Dual-Graph Initialization}
The Dual-Graph architecture comprises the GCG and the GGVG, each tailored to process specific encoded information types—geometric and visual.

To initialize the nodes of the Dual-Graph, We classify them into two main categories: ordinary matching points (P) and endpoints of matching line segments (L). These types of points are collectively referred to as Nodes (N) in the graph. This classification forms the basis for Dual-Graph analysis.

Following this classification, we introduce three key modules to initialize the node features in the Dual-Graph: Spatial Coordinate Encoder (SCE), Line Segment Encoder (LSE), and Pixel Encoder (PE). 

\subsubsection{Spatial Coordinate Encoder}
The SCE encodes the spatial relationships of all nodes (N) in the GCG. We concatenate the coordinates of the matching nodes along the coordinate dimension and use the SCE to convert these coordinates into high-dimensional space.

\begin{equation}
f^{0}_{i,N} = SCE([x_r||y_r||x_t||y_t]).\label{eq3}
\end{equation}

Here, $[...||...]$ represents concatenation.

Matched line segments are represented by their endpoints and are treated similarly to regular matched points during SCE encoding. This approach minimizes the loss of geometric information.

\subsubsection{Line Segment Encoder}
The LSE encodes the structured representation of line segments. The encoder takes the coordinates of the line segment endpoints (L) and their differences to capture the line segment's length and structure. Let's define the differences:
\begin{equation}
\Delta x_{r,L} = x_{r,L}^{'} - x_{r,L}, \quad \Delta y_{r,L} = y_{r,L}^{'} - y_{r,L}.
\end{equation}
\begin{equation}
\Delta x_{t,L} = x_{t,L}^{'} - x_{t,L}, \quad \Delta y_{t,L} = y_{t,L}^{'} - y_{t,L}.
\end{equation}
The initial line segment feature $f_{i,L}^{0}$ is then represented as:

\begin{equation}
    f_{i,L}^{0} = LSE([x_{r,L} || y_{r,L}|| \Delta x_{r,L} || \\
    \Delta y_{r,L} || x_{t,L}|| y_{t,L}|| \Delta x_{t,L} || \Delta y_{t,L}]).
\end{equation}

The LSE acts as a complement to SCE in the GCG by integrating geometric information from line segments with the topological structure of the GCG, effectively augmenting the structured information of the scene.

\subsubsection{Pixel Encoder}
We abandoned the traditional ResNet method for extracting visual features because it primarily focuses on high-level features, which may overlook the fine-grained spatial details necessary for precise pose estimation. Instead, we use the Pixel Encoder (PE) to initialize visual features in the GGVG. The PE encodes the pixel values sampled at all nodes in the GCG into high-dimensional visual features, preserving detailed spatial information for accurate pose estimation.
\begin{equation}
f^{0}_{i,V} = PE([pixel(x_{r},y_{r}),pixel(x_{t},y_{t})]).
\end{equation}
Here, $pixel(x_{i},y_{i})$ represents the sampled image pixel value based on the coordinates of all Nodes. 

% $f^{0}_{i,N}$, $f^{0}_{i,L}$, $ f ^{0}_{i,V}$ represent features of all Nodes(N), features of line segment (L) and the initial input feature of Geometry-Guided Visual Graph.

\subsubsection{Dual-Graph Update Rules}
To establish the overall node update rules for the Dual-Graph, we set three types of edges for node connections:
\begin{itemize}
    \item Intra-Graph Self-Attention Edge: Connects all nodes in the same graph.
    \item Line Segment Endpoint Edge: Connects the two endpoints of the same line-matching segment.
    \item Inter-Graph Cross-Attention Edge: Connects all nodes between GCG and GGVG.
\end{itemize}

With these foundational elements, we then elaborate on our Dual-Graph Architecture. 

% The innovative aspect of this architecture lies in its ability to dynamically fuse geometric and visual data through a weighted integration process, enabling the effective learning of pose estimations with enhanced accuracy and reliability. 

\subsection{Geometric Correspondence Graph}
In relative pose estimation tasks, geometric information is vital in describing scene structure and details, providing constraints for the final results. GCG makes full use of point and line segment features and updates geometric features in two stages according to the graph's topological structure. This method reflects the structured information of the scene and improves the model's understanding of geometric relationships.

\subsubsection{Intra-Graph Self-Attention Edge}
In the $l$-th layer of the GCG, the $i$-th node is represented as a high-dimensional feature $f^{l}_{i,N}$. The node updates its feature by aggregating messages from all other nodes in the GCG. The message propagation and feature update are defined as follows:
\begin{equation}
f^{l+1}_{i,N} = f^{l}_{i,N} + F_g([f^{l}_{i,N} || m^{l}_{i,N}]).
\end{equation}
Here, $f^{l+1}_{i,N}$ is the updated node feature in the $l+1$-th layer, $[...||...]$ represents concatenation, and $F_g$ represents an MLP. The Message $m^{l}_{i,N}$ is aggregated through multi-head self-attention:
\begin{equation}
m^{l}_{i,N} = \sum_{j} softmax\left(\frac{Q^{l}_{i,N} {K^{l}_{j,N}}^{T}}{\sqrt{D_k}}\right) V^{l}_{j,N} \label{eq4}
\end{equation}

In this context:

\begin{itemize}
    \item $Q^{l}_{i,N}$ represents the query vector for node $i$ at the $l$-th layer.
    \item $K^{l}_{j,N}$ and $V^{l}_{j,N}$ denote the key and value vectors for node $j$ at the $l$-th layer.
    \item $D_k$ is the dimension of the key vectors used for scaling the softmax computation.
\end{itemize}

The self-attention mechanism calculates the attention scores between node $i$ and all other nodes $j$ by taking the dot product of $Q^{l}_{i,N}$ and $K^{l}_{j,N}$, normalizing by $D_k$, and applying the softmax function to determine the relative importance of each node $j$ to node $i$.

\subsubsection{Line Segment Endpoint Edge}
In the message passing of the GCG, we integrate structural information from line segments to aid pose estimation. As shown in the right half of the GCG section in Fig. \ref{Fig1}, this process involves aggregating the features of nodes classified as line segment endpoints (L) along with their corresponding endpoint encodings, defined as:

\begin{equation}
f^{l+1}_{i,L} = f^{l}_{i,L} + \sum_{j}\frac{F_L\left([f^{l}_{i,N} || f^{l}_{j,N} || f^{0}_{j,L}]\right)}{N_j}.
\end{equation}

Here, $F_L$ represents an MLP, $f^{l}_{i,N}$ and $ f^{l}_{j,N}$ represents the feature of two endpoints of a line segment, $f^{0}_{j,L}$ represents the line segment feature obtained by LSE, and $N_j$ represents the number of neighbors of node $i$.

% This aggregation is crucial for message propagation in the graph. By incorporating these line segmental structures, we significantly enhance the graph's ability to understand and interpret complex spatial relationships and geometrical information in the scene. Line structures, with their defined length and orientation, offer a richer context than isolated points, enabling a more robust and nuanced understanding of the scene's geometry. Line segments can provide valuable constraints and additional information that are not available from points alone.

% For instance, the alignment and relative positioning of these line segments can offer insights into the perspective and orientation of objects in the image, thereby improving the pose estimation performance. Furthermore, line segments can serve as strong indicators of structural regularities and symmetries in the scene, which are key to resolving ambiguities in pose determination.

% Therefore, our Line Segment Structural Integration Message Passing process facilitates the integration of line segmental information and leverages unique geometric insights. This results in a more accurate, reliable, and geometrically informed estimation of pose, significantly advancing the capabilities of our graph-based approach in complex visual environments.

Integrating these line segment structures enhances the GCG's ability to interpret complex spatial relationships and geometrical information. Line segments provide valuable constraints and additional information that cannot be obtained from points alone. They indicate structural regularities and symmetries within the scene, improving pose estimation.

Thus, the Line Segment Structural Integration message-passing process leverages geometric insights, resulting in more accurate and reliable pose estimation in complex visual environments.

\subsection{Geometry-Guided Visual Graph}
In the relative pose estimation task, visual information can provide fine-grained details of the scene, which is critical for improving the model's accuracy. Compared to the high-level features extracted by ResNet, fine-grained image information captures important scene details more comprehensively. Convolutional neural networks such as ResNet are good at extracting high-level features but may overlook key spatial details, especially in complex scenes. Fine-grained image information preserves local subtle changes and the global structure, which is essential for accurate pose estimation.

The GGVG samples image information by using the coordinates of all nodes in GCG to obtain spatial details. This method preserves essential spatial information and accurately captures minor differences in the two frames. For example, this information can help models recognize minor street signs and building edges in complex urban environments.

\subsubsection{Intra-Graph Self-Attention Edge}
Similarly, in the $l$-th layer of GGVG, the $i$-th node is represented as a high-dimensional feature $f^{l}_{i,V}$. The feature update through message propagation is defined as:
\begin{equation}
f^{l+1}_{i,V} = f^{l}_{i,V} + F_v([f^{l}_{i,V} || m^{l}_{i,V}]).
\end{equation}
Here, $F_v$ represents an MLP. The message $m^{l}_{i,V}$ is aggregated through multi-head self-attention:
\begin{equation}
m^{l}_{i,V} = \sum_{j} softmax\left(\frac{Q^{l}_{i,V} {K^{l}_{j,V}}^{T}}{\sqrt{D_k}}\right) V^{l}_{j,V}.
\end{equation}

The subscript $V$ represents the node in GGVG, and the meaning of the other symbols is the same as in GCG.

\subsubsection{Inter-Graph Cross-Attention Edge}
The GGVG has a close interaction between visual and geometric features. After nodes in the GGVG perform self-attention updates, they are guided by the features extracted from the GCG, forming a mutual relationship between geometry and visual appearance. The specific definition is as follows:
\begin{equation}
f^{l}_{i,V} = f^{l}_{i,V} + F_{guide}([f^{l}_{i,V} || m^{l}_{i,GG}]).\label{eq11}
\end{equation}

Here, $m^{l}_{i,GG}$ represents the Geometry-Guided message: 

\begin{equation}
m^{l}_{i,GG} = \sum_{j} softmax\left(\frac{Q^{l}_{i,G} {K^{l}_{j,V}}^{T}}{\sqrt{D_k}}\right) V^{l}_{j,V} \label{eq12}
\end{equation}

where $Q^{l}_{i,G}$ represents the output features of the $l$-th layer of the GCG, $K^{l}_{j,V}$, $V^{l}_{j,V}$represents the output features of the $l$-th layer of the GGVG,

The GGVG uses geometric features from each layer of the GCG as queries and the corresponding visual features as keys and values for cross-attention computation. This ensures that geometric context is integral to the updating process, enhancing interaction between geometric structures and visual cues.

% By incorporating geometric features as queries, the model retains more image details and improves the precision of visual feature updates. This leads to a more detailed and accurate understanding of the scene.

\subsection{Feature Weighted Fusion Module}
As mentioned in Zhuang et al. \cite{r14}, matching points with different spatial distances have varying impacts on the relative pose estimation task. In the graph, nodes with closer spatial distances require more attention for pose learning, which is less helpful for pose learning, so it is unfair to consider the contribution of each node feature in the graph to the final result on average. We must consider the contribution of each node feature in each graph in a different way. 

Inspired by \cite{r39}, we process the geometric features of the GCG and the visual features from the GGCG in our feature-weighted fusion module. As shown in the upper right part of Fig. \ref{Fig1}, both sets of characteristics undergo an average pooling layer to produce refined geometric and visual attributes. 

Next, we employ a Sigmoid function to determine the weight of each node in the Dual-Graph for the final pose estimation result. The weighted geometric and visual features are then concatenated and fed into two MLPs: one for relative pose estimation and the other for uncertainty determination. These MLPs generate the relative pose and the inverse variance between the image pairs $A$ and $B$.

The inverse variance serves as a measure of confidence in the estimated pose. This allows the model to account for varying levels of confidence in different parts of the image, enhancing the overall robustness and precision of the relative pose estimation. Higher inverse variance indicates a lower uncertainty, leading to more confident and accurate pose estimation. This allows the model to account for the varying confidence levels in different parts of the image, improving the overall robustness and precision of the relative pose estimation.

\subsection{Loss}
In the context of integrating Aleatoric Uncertainty \cite{r56} for relative pose estimation, our approach is different from the traditional loss, favoring a formulation that captures the inherent uncertainty in data. This loss is achieved by predicting both the expected value and the variance of our pose estimates, allowing for a representation of confidence in these estimations. Specifically, our loss function balances the fidelity of pose predictions with the model's certainty, expressed as: 
\begin{equation}
L = \log(\sigma^2) + \frac{(\theta - \mu)^2}{\sigma^2}
\end{equation}
where \(\theta\) is the Ground Truth (GT), \(\mu\) represents the model's pose predictions, and \(\sigma^2\) represents the predicted uncertainty. This loss allows for a more sophisticated assessment of the model's performance by considering the accuracy of the pose estimations and the confidence level associated with each prediction. This enables a nuanced optimization that prioritizes reliable data ambiguity and noise predictions.

\section{Experiments}

\label{section4}
\subsection{Dataset}
% \subsubsection{Dataset}
\textbf{DeMoN.} This dataset is composed of a substantial amount of real, synthetic, indoor, and outdoor data from MVS \cite{r53}, SUN3D \cite{r52}, RGB-D \cite{r54}, and Scenes11 \cite{r55}, encompassing diverse scene textures and geometric information. Each dataset provides RGB image sequences, camera intrinsics, and camera poses. We adopt the same training and testing set split as DeMoN \cite{r10} and follow DeepSfM \cite{r11}'s data preparation standards. 

\textbf{KITTI Odometry.} This dataset offers a substantial collection of camera relative pose estimation data from natural-world driving sequences. In line with previous work, we employ Seq.00-Seq.08 for training and Seq.09-Seq.10 for testing, respectively.

\subsection{Metrics}
We employ the popular evaluation metrics used in prior work to compare our method with previous approaches in relative pose estimation.

In the DeMoN \cite{r10} dataset, the error between the predicted relative rotation and translation and GT is defined in terms of angular difference measured in degrees, as follows:
\begin{equation}
E_{rot}^{demon}=\arccos{(q_{gt}\cdot q_{pred})}
\end{equation}
\begin{equation}
E_{tran}^{demon}=\arccos{(t_{gt}\cdot t_{pred})}
\end{equation}
where $q_{pred}$ and $t_{pred}$ represent the rotation and translation of the network output, which are represented by quaternions and normalized translation, respectively, while the $Q_{gt}$ and $t_{gt}$ represent the ground truth poses.

In the KITTI Odometry \cite{r24} dataset, we utilize the standard Odometry evaluation metrics, namely the translation error ($t_{rel}$) and rotational error ($r_{rel}$). The formula is as follows:
\begin{equation}
\Delta P_{gt}=P_{gt,A}^{-1} P_{gt,B} 
\end{equation}
\begin{equation}
\Delta P_{pred}=P_{est,A}^{-1} P_{est,B}
\end{equation}
\begin{equation}
E^{kitti}=\Delta P_{est}^{-1} \Delta P_{gt}
\end{equation}
\begin{equation}
E_{rot}^{kitti}= \left( \frac{1}{m} \sum_{i=1}^{m} \left\| \text{rot}(E_i) \right\|^2 \right)^{\frac{1}{2}}
\end{equation}
\begin{equation}
E_{rot}^{kitti}= \left( \frac{1}{m} \sum_{i=1}^{m} \left\| \text{tran}(E_i) \right\|^2 \right)^{\frac{1}{2}}
\end{equation}

For each pair of images $A$,$B$, we calculate the relative change in pose $\Delta P_{gt}$ between the ground-truth poses and $\Delta P_{pred}$ between the estimated poses. Then, we calculate the pose error $E^{kitti}$ for each pair of time points. Finally, we obtain an overall value for rotation and translation parts using the Root Mean Square Error (RMSE).

\subsection{Implement Details}
Our network is implemented in the PyTorch framework. During training, matching points and line segments are extracted from images with a resolution set at 192x256 in the DeMoN \cite{r10} datasets, and the resolution is set at 128x416 in the KITTI Odometry datasets \cite{r24}. The batch size is set to 16. To ensure the smooth training of the model, we adopted a strategy similar to  UA-Fusion \cite{r14} by randomly repeating and deleting the obtained matching results, resulting in 384 matching points and 192 matching line segments on different image pairs. Training is carried out using the Adam optimizer with a learning rate of $1 \times 10^{-4}$. 

\def\tablename{\large Table}
\begin{table*}[ht]\centering
\renewcommand{\arraystretch}{1.2}
\setlength{\tabcolsep}{3pt}{
\caption{Comparison with the state-of-the-art on DeMoN \cite{r10} datasets. The best results are in bold and the second-best results are in underlined.}
\label{table1}
% \refstepcounter{table}
\begin{tabular}{|c|cc|cc|cc|cc|cc|}
\hline
\multirow{2}{*}{} & \multicolumn{2}{c|}{MVS} & \multicolumn{2}{c|}{Scenes1l} & \multicolumn{2}{c|}{RGB-D} & \multicolumn{2}{c|}{Sun3D} & \multicolumn{2}{c|}{All} \\ \cline{2-11} 
                  & Rot.        & Tran.      & Rot.          & Tran.         & Rot.        & Tran.        & Rot.        & Tran.        & Rot.       & Tran.      \\ \hline
DeMoN \cite{r10}        & 5.156                & 14.447               & 0.809                & 8.918                & 2.641                & 20.585               & 1.801                & 18.811                &2.258                &14.821     \\
LS-Net \cite{r12}        & 4.653                & 11.221               & 4.653                & 8.210                & \textbf{1.010}                & 22.110               & 1.521                & 14.347                 &3.095                &13.300\\
BA-Net \cite{r13}       & 3.499                & 11.238               & 3.499                & 10.370               & 2.459                & 14.900               & 1.729                & 13.260                &3.190             &12.887                    \\
DeepSfM \cite{r11}      & 2.824               & 9.881                & \textbf{0.403}                & 5.828                & 1.862                & 14.570               & 1.704                & 13.107                &1.483                &12.064                 \\
LGC-Net \cite{r40}      & \textbf{2.753}                & \textbf{3.548}               & 0.977                & \underline{4.861}                 & 2.014                & 16.426               & \textbf{1.386}                & 14.118                &1.646                &\underline{9.322}                 \\
NM-Net \cite{r41}      & 6.628                & 12.595               & 15.717               & 31.477               & 13.444               & 34.212               & 4.393                & 21.091                &8.893                &26.294                 \\
DeepSfM+RUMs \cite{r66}         & 2.851                & 10.034                & \underline{0.445}                & 4.902                & 1.606                & \underline{12.741}               & 1.520                & \textbf{12.586}                &\underline{1.399}                &9.367                 \\ \hline
Our        & \underline{2.818} & \underline{5.143}            & 0.479  & \textbf{3.523}                & \underline{1.494}  & \textbf{12.533}            & \underline{1.511} & \underline{12.739}                &\textbf{1.378}                &\textbf{7.944}                        \\ \hline
\end{tabular}
}
\end{table*}

\subsection{Results on DeMoN}
% \textbf{Comparison:} On the DeMoN \cite{r10} dataset, we compared our method with state-of-the-art relative pose estimation methods using depth constraints, as shown in Table \ref{table1}, including DeMoN \cite{r10}, LS-Net \cite{r12}, BA-Net \cite{r13}, DeepSfM \cite{r11}, LGC-Net \cite{r40} and NM-Net \cite{r41}. Among these, DeMoN \cite{r10} employs optical flow and depth constraints to iteratively predict camera poses. LS-Net \cite{r12} uses a Long Short-Term Memory Regression Neural Network (LSTM-RNN) to iteratively refines depth and pose estimation. BA-Net adjusts the optimization of scene depth and camera motion through feature metric Bundle Adjustment, and the entire process is differentiable. DeepSfM \cite{r11} incorporates depth disparity into the cost volume, further strengthening the geometric consistency between camera poses and their current depth estimates. DeepSFM + RUMs} \cite{r66} adds a residual motion depth update module to DeepSFM} \cite{r11}, which can better handle dynamic scenes}. It is worth noting that both BA-Net \cite{r13} and DeepSfM \cite{r11} serve as post-processing networks for DeMoN \cite{r10} to optimize poses.

On the DeMoN \cite{r10} dataset, we conducted a comprehensive comparison of our method with state-of-the-art relative pose estimation methods using depth constraints, including DeMoN \cite{r10}, LS-Net \cite{r12}, BA-Net \cite{r13}, DeepSfM \cite{r11}, LGC-Net \cite{r40}, NM-Net \cite{r41} and DeepSfM + RUMs \cite{r66}, as shown in Table \ref{table1}. LGC-Net \cite{r40} and NM-Net \cite{r41} are networks designed to learn matching points, and the final poses are obtained through 5-point methods \cite{r5} and Bundle Adjustment \cite{r7}. 

The quantitative comparisons are shown in Table \ref{table1}. Our method produces competitive results compared to methods that use depth or initial pose as a constraint.

% This success is due to the incorporation of line segment geometric structure representation in our network, which effectively handles environments with low-texture and repetitive-texture scenarios, both in synthetic and real-world settings.

The Scenes11 dataset contains many artificially synthesized scenes, which often have complex geometric structures and neat object arrangements, and the ground truth has high authenticity. Our method accurately captures the geometry information of the line segments in the scene. This line segment geometry is effective in capturing the translation information of the scene because it better describes the geometric layout. However, rotation estimation often needs to capture more subtle changes in relative posture, which may appear as small angle differences in the image. In the Line Segment Encoder, we encode more line segment length information than angle information, resulting in a poorer representation in rotation estimation than the method using depth constraints. Methods based on depth difference optimization, such as DeepSfM \cite{r11} perform well in rotation estimation because the depth difference is directly related to the rotation of the camera. Depth information can reflect the three-dimensional structure of the scene in more detail. When the scene rotates, the change in depth can be directly fed back to the rotation estimation, so these methods often have smaller rotation errors.
The MVS dataset has the characteristics of multi-view views, and the scene changes are significant and may include different perspectives and occlusions. The Sun3D dataset includes many dynamic and complex indoor scenes, which usually contain many occluders and fast-moving objects. In this case, the detection of line segment geometry may have certain problems. By learning and extracting more robust feature points, LGC-Net \cite{r40} can find consistent features in multi-view data, thereby effectively dealing with perspective changes and occlusion problems. After completing feature point matching, LGC-Net \cite{r40} uses the classic 5-point method and bundle adjustment for pose estimation. However, our method still achieves suboptimal translation error on the Sun3D dataset, which is only 1.2\% higher than the DeepSFM + RUMs \cite{r66} result, thanks to a certain amount of line segment geometry information.

Despite the advantages of using structured line segments, the reliance of our method solely on 2D image cues presents a limitation compared to methods that utilize comprehensive 3D data. In particular, our approach relies on the performance of upstream line segment detection. However, as seen in the KITTI Odometry \cite{r24} datasets in the next section, our method performs well in natural road scenes filled with many line segment structures. In the natural road scene of KITTI, the performance of line segment detection is more robust. This showcases a promising direction for pose estimation in scenarios where 3D data may be unavailable, emphasizing the potential of our structured line segment approach.

\subsection{Results on KITTI Odometry}
% On the KITTI Odometry \cite{r24} datasets, Str-L Pose successfully employs 2D image information to estimate relative poses, offering a practical alternative to 3D data-dependent methods. Our approach, effective in real-world driving sequences, highlights the potential of 2D data for complex environments. The quantitative and qualitative results testify our method's capability.
On the KITTI Odometry \cite{r24} dataset, we quantitatively compare with several state-of-the-art methods using depth constraints. We also plot the trajectories using our method on the 09,10 sequence to demonstrate our better results.
\begin{table}[ht]
\centering
\renewcommand{\arraystretch}{1.2}
\caption{Comparison with depth constrained methods on Seq.09 And Seq.10 of the KITTI Odometry datasets. $t_{rel}$: average translation RMSE Drift(\%); $r_{rel}$: average rotation RMSE Drift($\circ/100m$). The best performance is in bold, and the second best is underlined.}
\label{table2}
\begin{tabular}{|c|cc|cc|}
\hline
\multirow{2}{*}{Methods}                     & \multicolumn{2}{c|}{Seq.09}          & \multicolumn{2}{c|}{Seq.10} \\ \cline{2-5} 
                                             & $t_{rel}$ & \multicolumn{1}{c|}{$r_{rel}$} & $t_{rel}$       & $r_{rel}$       \\ \hline
% DSO \cite{r1}               & 28.1   & \textbf{0.21}                        & 24.0         & \textbf{0.22}         \\
% ORB-SLAM2(w/o LC) \cite{r6} & 9.31   & 0.26                        & \textbf{2.66}         & 0.39         \\
% ORB-SLAM2 \cite{r6}         & \textbf{2.84}   & 0.25                        & 2.67         & 0.38         \\ \hline
SfMLearner \cite{r22}       & 11.34  & 4.08                        & 15.26        & 4.08         \\
GeoNet \cite{r21}           & 21.81  & 8.12                        & 19.43        & 7.32         \\
UnDeepVO \cite{r42}         & \underline{7.01}   & 3.60                        & 10.63        & 4.65         \\
Deep-VO-Feat \cite{r43}     & 9.55   & 3.90                        & 9.79         & 3.44         \\
Wang et al. \cite{r44}      & 9.88   & 3.40                        & 12.24        & 5.20         \\
MonoDepth2 \cite{r45}       & 10.85  & 2.86                        & 11.60        & 5.72         \\
SC-SfMLearner \cite{r46}    & 11.2   & 3.35                        & 10.1         & 4.96         \\
CC \cite{r47}               & 7.57   & \underline{1.84}                        & 8.10         & 3.38         \\
FeatDepth \cite{r48}        & 8.75   & 2.11                       & 10.67        & 4.91         \\
Mai et al. \cite{r49}       & 7.62   & 2.53                        & 10.38        & 2.99         \\
Lee et al. \cite{r50}       & 8.6    & 2.9                         & 9.2          & 4.5          \\
Liang et al. \cite{r51}     & 8.13   & 2.95                        & \underline{6.76}         & \underline{2.42}         \\ \hline
% Zhou et al. \cite{r67}     & 4.24   & 1.49                        & 5.29         & 2.34         \\ \hline
XVO \cite{r67}     & -   & -                        & 12.17         & 3.45         \\
SCIPaD \cite{r68}     & 7.43   & 2.46                        & 9.82         & 3.87         \\
Ours                                         & \textbf{5.03}   & \textbf{1.57}                       & \textbf{4.40}         & \textbf{2.05}         \\ \hline
\end{tabular}
\end{table}

\textbf{Quantitative Experiments: }
 % several geometry-based methods: DSO \cite{r1}, ORB-SLAM2 \cite{r6} and
In quantitative terms, we compared our method against several state-of-the-art depth-constrained relative pose estimation methods: SfMLearner \cite{r22}, GeoNet \cite{r26}, UnDeepVO \cite{r42}, Deep-VO-Feat \cite{r43}, Wang et al.\cite{r44}, MonoDepth2 \cite{r45}, SC-SfMLearner \cite{r46}, CC \cite{r47}, FeatDepth \cite{r48}, Mai et al. \cite{r49}, Lee et.al. \cite{r50}, Liang et.al. \cite{r51}, XVO \cite{r67}, and SCIPaD \cite{r68}, as shown in Table \ref{table2}. This evaluation employed standard odometry metrics, including translation and rotation errors, to measure the accuracy of pose estimations across different sequences in the KITTI Odometry \cite{r24} datasets. 

% Our model performs well when compared to the methods that use depth information for pose estimation constraints. Especially compared to the DeMon datasets \cite{r10}, our method has made significant progress, mainly attributed to the KITTI Odometry \cite{r24} datasets being a real road scene with a wide range of structured line segment information, particularly in Seq.10 characterized by challenging structural features. This demonstrates that leveraging structured line segments and Dual-Graph is effective in real environments.

Our model shows better performance compared to methods that use depth information for pose estimation constraints. Specifically, on the KITTI Odometry \cite{r24} datasets, particularly in Seq.10, our method demonstrates the effectiveness of leveraging structured line segments in natrual road scenes with complex structural features.

\textbf{Qualitative Experiments:}
\begin{figure}
\centering
\includegraphics[width=\linewidth]{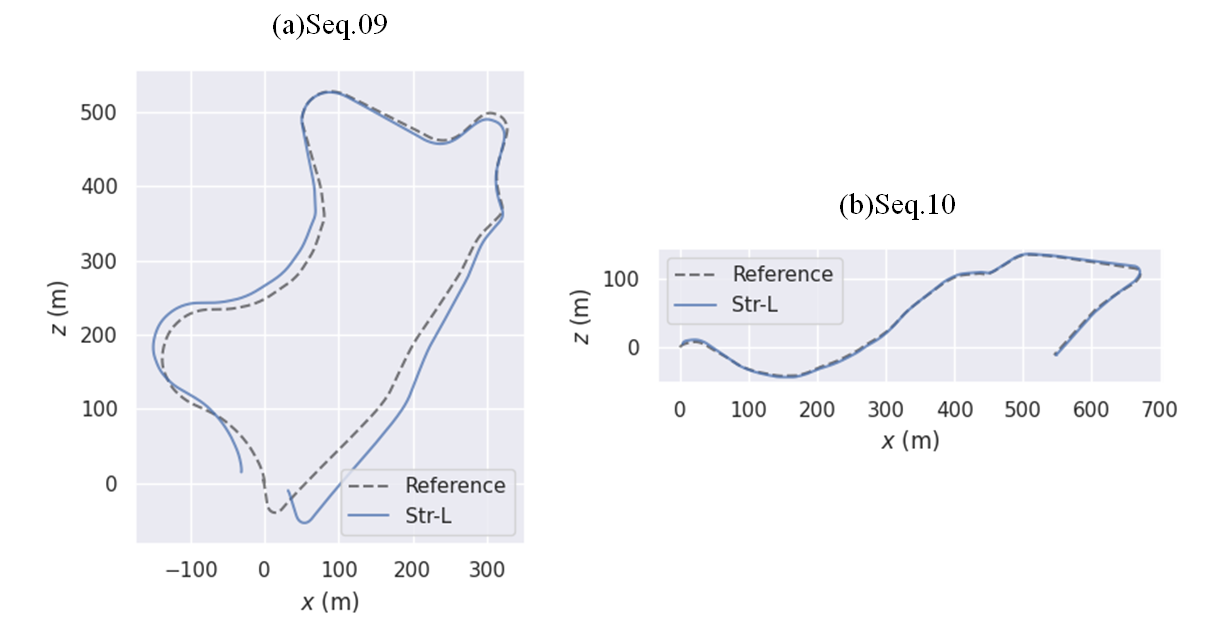}
\caption{The trajectory using our method on (a) Seq. 09 and (b) Seq. 10 of KITTI Odometry datasets.}
\label{fig2}
\end{figure}
We have charted the trajectories generated by the predicted relative poses of these methods, as shown in Fig.\ref{fig2} , offering a visual testament to our model's performance. In Seq.10, characterized by structured environments with many buildings, our method is almost consistent with the ground truth. This demonstrates the effectiveness of our method.
% This remarkable achievement is largely attributed to the novel application of structured line segments in our network's architecture.

\begin{figure}
\centering
\includegraphics[width=\linewidth]{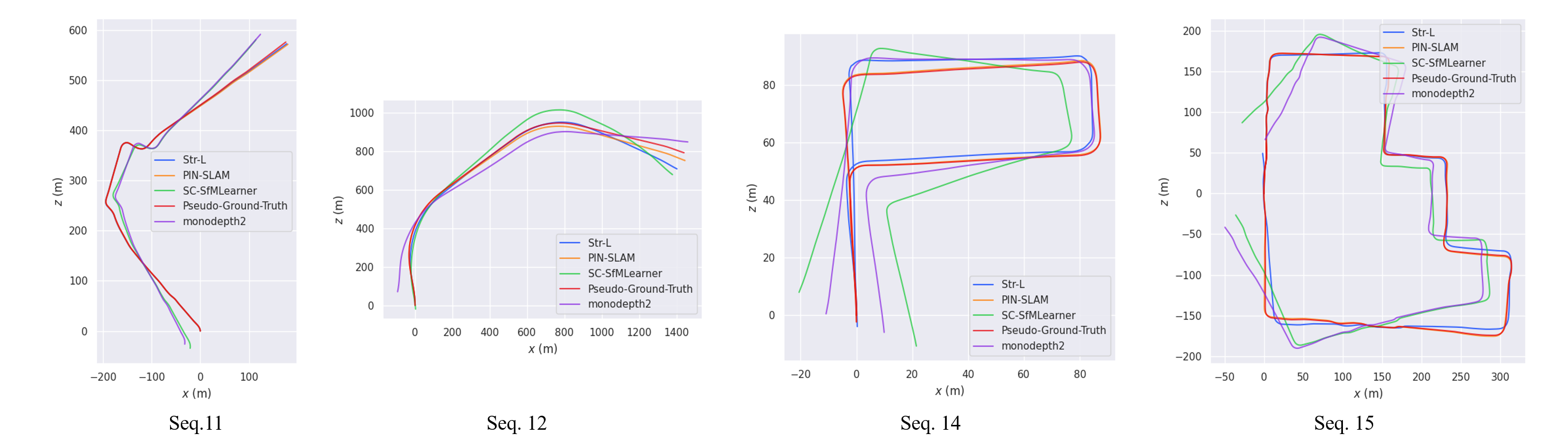}
\caption{Comparison of trajectories of our method with the depth-constrained methods and other high-precision LiDAR SLAM methods on Seq. 11, Seq 12, Seq. 14 and Seq. 15 of the KITTI odometry dataset.}
\label{fig3}
\end{figure}

To further confirm the effectiveness of our method, we selected four additional KITTI Odometry sequences. Two sequences have more structured scenes of buildings (Seq.11, Seq.15), and the other two sequences contain more natural environments (Seq.12, Seq.14). Since these four sequences do not provide pose ground truth, we use the F-LOAM \cite{r69} algorithm to estimate the pseudo ground truth of the pose. F-LOAM is a high-precision LiDAR SLAM algorithm with a small error on the KITTI Odometry Benchmark. We compare our method with two other depth-constrained methods. Finally, we add another LiDAR SLAM algorithm, PIN-SLAM,  \cite{r70} to supervise the pseudo ground truth's effectiveness and verify the method's performance. The trajectory is shown in Fig. \ref{fig3} .

Due to the dense arrangement of buildings and significant structural features in Seq. 11 and Seq. 15, our method can accurately capture these features and show high stability and accuracy in trajectory estimation. It is even almost consistent with the pseudo ground truth and the trajectory predicted by the LiDAR SLAM algorithm. Our method shows higher accuracy, especially in the processing of long straight sections and intersections.

In Seq.12 and Seq.14, the natural environment in these two sequences includes more unstructured features, such as large sky areas. Although the geometric structure is relatively sparse in such scenes, our method can still estimate the vehicle trajectory relatively stably, outperforming methods using depth constraints.

\begin{table}[ht]
\centering
\renewcommand{\arraystretch}{1.2}
\setlength{\tabcolsep}{4pt}{
\caption{Ablation Study On DeMoN \cite{r10} Datasets. The Best Results Are In Bold.}
\label{table3}
\begin{tabular}{|c|cc|cc|cc|cc|cc|}
\hline
\multirow{2}{*}{} & \multicolumn{2}{c|}{MVS} & \multicolumn{2}{c|}{Scenes1l} & \multicolumn{2}{c|}{RGB-D} & \multicolumn{2}{c|}{Sun3D} & \multicolumn{2}{c|}{All} \\ \cline{2-11} 
                  & Rot.        & Tran.      & Rot.          & Tran.         & Rot.        & Tran.        & Rot.        & Tran.        & Rot.       & Tran.       \\ \hline
Baseline          & 3.558       & 6.445      & 0.580         & 4.375         & 2.043       & 18.184       & 2.003       & 18.811       & 1.787      & 9.742       \\
Baseline + L      & 3.777       & 6.834      & 0.592         & 5.272         & 2.030       & 17.096       & 1.928       & 14.858       & 1.813      & 10.402      \\
Baseline + LP     & 3.176       & 6.054      & 0.543         & 4.335         & 1.815       & 14.480       & 1.643       & 12.401       & 1.570      & 8.771       \\
Baseline + R      & 3.116       & 5.972      & 0.541         & 4.258         & 1.675       & 14.679       & 2.042       & 13.890       & 1.617      & 9.109       \\
Baseline + V      & 3.129       & 6.064      & 0.525         & 4.798         & 1.797       & \textbf{10.489}       & 1.800       & 13.399       & 1.586      & 8.264       \\
Baseline + V + W  & 3.162       & 6.074      & \textbf{0.473}         & 3.912         & 1.853       & 11.953       & 1.678       & 12.797       & 1.559      & 8.140       \\
Baseline + LP + R & 3.219       & \textbf{4.298}      & 0.758         & 4.130         & 1.778       & 17.805       & 1.737       & 12.380       & 1.668      & 9.116       \\
Baseline + LP + V & 3.232       & 5.720      & 0.539         & 4.255         & 2.165       & 14.752       & 1.634       & \textbf{12.321}       & 1.656      & 8.723       \\ \hline
Full Model        & \textbf{2.818}       & 5.143      & 0.479         & \textbf{3.523}         & \textbf{1.494}       & 12.533       & \textbf{1.511}       & 12.739       & \textbf{1.378}      & \textbf{7.944}       \\ \hline
\end{tabular}
}
\end{table}

\subsection{Ablation Studies}
% To demonstrate the impact of each proposed component in our Str-L Pose framework, we conducted ablation studies on the DeMoN \cite{r10} test dataset in Table \ref{table3}. 
To demonstrate the impact of each proposed component in our Str-L Pose framework, we perform an ablation study using all test data from DeMoN \cite{r10}. The results are shown in Table \ref{table3}. These experiments dissect the contributions of individual elements to the overall performance enhancement in relative pose estimation. 

Our ablation framework includes several key variants of the proposed model, each omitting a specific component or integrating alternative methodologies for comparison. The variants are as follows:

\begin{itemize}
    \item \textbf{Baseline}: Utilizes only matching points for pose estimation, excluding advanced graph feature extraction.
    \item \textbf{Baseline + Lines (Baseline + L)}: Integrates matching line segments with the baseline model, including nodes represented by both matching points and endpoints of line segments.
    \item \textbf{Baseline + Line Segment Structural Integration Message Passing (Baseline + LP)}: Incorporates the Line Segment Structural Integration Message Passing module.
    \item \textbf{Baseline + ResNet (Baseline + R)}: Employs ResNet-34 for image feature extraction.
    \item \textbf{Baseline + Geometry-Guided Visual Graph (Baseline + V)}: Integrates the Geometry-Guided Visual Graph module.
    \item \textbf{Baseline + LP + R}: Combines Line Segment Structural Integration Message Passing and ResNet-34 with the baseline.
    \item \textbf{Baseline + LP + V}: Combines Line Segment Structural Integration Message Passing and Geometry-Guided Visual Graph with the baseline.
    \item \textbf{Baseline + V + W}: Combines Geometry-Guided Visual Graph and Feature Weighted Fusion Module with the baseline.
    \item \textbf{Full Model (Baseline + V + LP + W)}: Integrates all proposed components, representing the complete Str-L Pose framework.
\end{itemize}

\begin{table}[ht]
\centering
\renewcommand{\arraystretch}{1.2}

\caption{Results of a single image pair in the RGB-D subset. The first four results show that the baseline + V translation results are better than the Full Model, and the last four show that the B+V translation results are worse than the Full Model. The better translation results are in bold.}
\setlength{\tabcolsep}{4pt}{
\label{table4}
\begin{tabular}{|c|cc|cc|}
\hline
                  & \multicolumn{2}{c|}{Baseline + V} & \multicolumn{2}{c|}{Full Model} \\ \hline
Image Pair Number & Rot.        & Tran.               & Rot.      & Tran.               \\ \hline
08                & 1.823       & \textbf{0.314}      & 0.860     & 0.558               \\
38                & 0.660       & \textbf{1.149}      & 0.221     & 4.661               \\
46                & 1.914       & \textbf{1.388}      & 1.686     & 2.568               \\
78                & 1.158       & \textbf{2.060}      & 0.383     & 3.734               \\ \hline
04                & 1.321       & 15.222              & 1.677     & \textbf{11.137}     \\
12                & 0.703       & 8.334               & 0.451     & \textbf{3.765}      \\
22                & 0.442       & 8.024               & 0.419     & \textbf{5.375}      \\
31                & 0.285       & 3.975               & 0.446     & \textbf{1.732}      \\ \hline
\end{tabular}
}
\end{table}

\begin{figure}
\centering
\includegraphics[width=\linewidth]{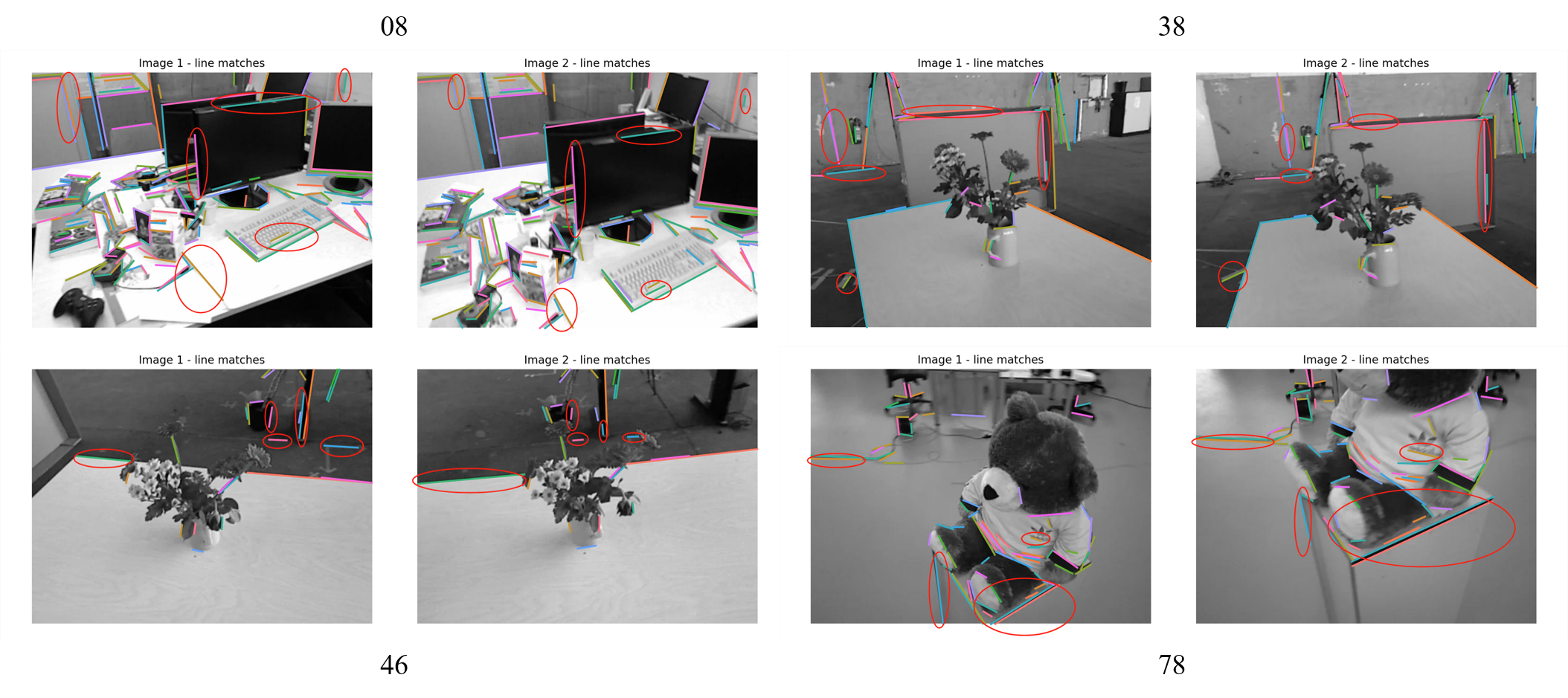}
\caption{Visualization of line segment matching results for image pairs whose Full Model translation results is worse than Baseline + V on the RGB-D subset. The red circles indicate matching line segments with obvious errors in line segment length.}
\label{fig4}
\end{figure}

\begin{figure}
\centering
\includegraphics[width=\linewidth]{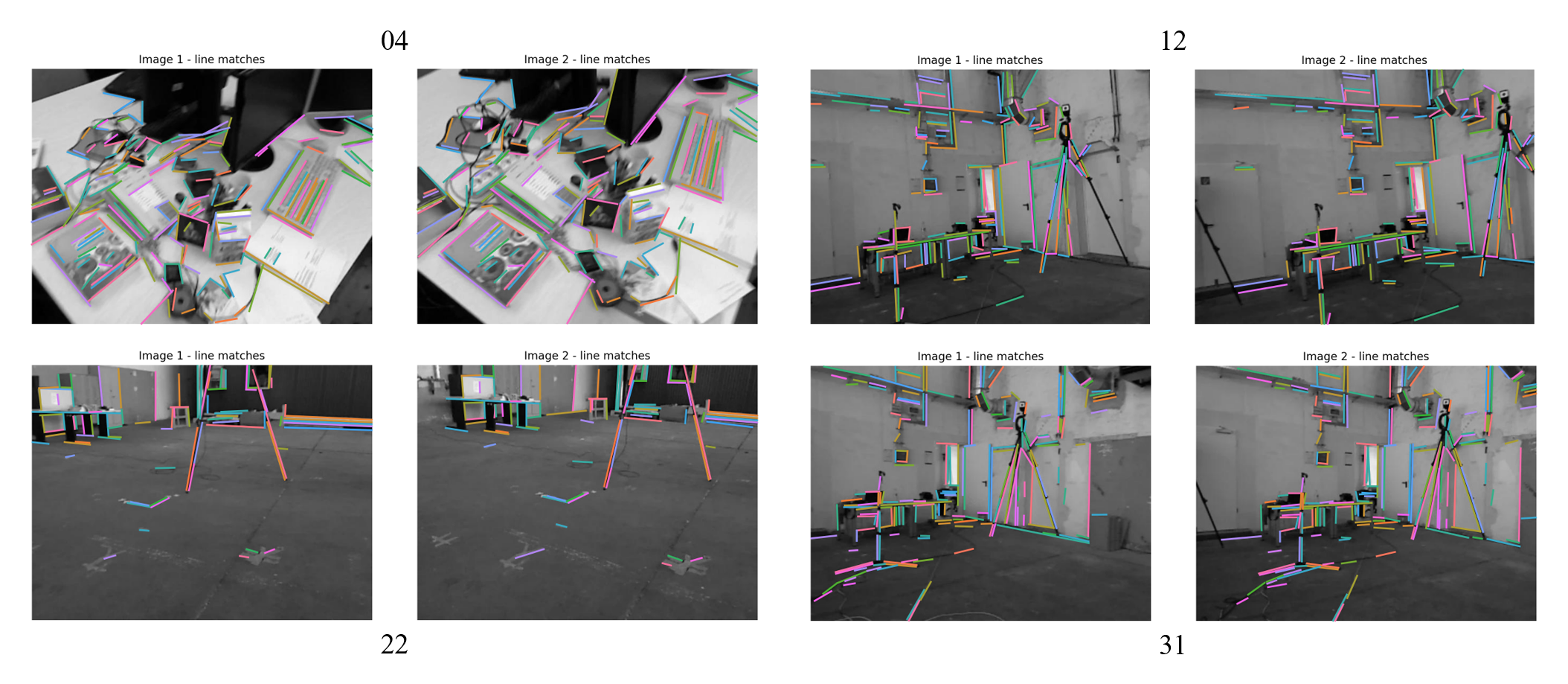}
\caption{Visualization of line segment matching results of image pairs where the Full Model translation results is better than Baseline + V on the RGB-D subset.}
\label{fig5}
\end{figure}

\textbf{Results and Discussion: }Key insights from the ablation studies include:
\begin{enumerate}
    \item \textbf{Analysis of Baseline and Baseline + L Performance:} Baseline + L exhibits slightly higher errors on the DeMoN \cite{r10} dataset than the Baseline. It is likely due to the duplicates among some matched line segments and matched points in image pairs. According to \cite{r14}, points farther apart provide stronger signals for motion estimation by leveraging the perspective effects in the field of view. In the Baseline + L experiment, the endpoints from matched line segments increase the likelihood of having more closely spaced matched points. This alters the original distribution of matched points, leading to a degradation in performance.
    \item \textbf{Comparative evaluation of module integration:} The evaluation of each proposed module in isolation—specifically, Baseline vs. Baseline + LP, Baseline + R vs. Baseline + V, and Baseline + V vs. Baseline + V + W—reveals incremental advancements in accuracy. Experimental results underscores the unique contribution of each component.
    \item \textbf{Discussion of DeMoN Subset Results:} Detailed analysis revealed that specific subsets, particularly the RGB-D dataset, presented unique challenges. For instance, the Full Model showed a higher translation error than the Baseline + V. We found that significant depth variations within the RGB-D subset can lead to increased occlusion, making line segment detection less accurate, especially for endpoint detection. This can be seen in the visualization of the line segment matching in Fig \ref{fig4} and Fig. \ref{fig5}, where Fig. \ref{fig4} has more matching line segments of incorrect length than Fig. \ref{fig5} . Table \ref{table4} shows the rotation and error results of the image pairs in Fig. \ref{fig4} and \ref{fig5}. If the line segment lengths in the left and right views do not match, the object is projected at different scales in the two views, indicating inconsistent scaling or perspective distortion. This distortion affects the accurate estimation of the translation vector, as translation estimation requires the objects in the two views to have the same projection length relationship.
    
    However, rotation estimation is less affected by these wrong line segment lengths. It is because that rotation estimation depends primarily on the relative orientation and angle between features rather than their exact spatial locations. Therefore, the Full Model can achieve the best rotation accuracy on the RGB-D subset.
    \item \textbf{Superiority of Full Model:} The performance improvement brought by module combinations (Baseline + LP + V, Baseline + V + W) is not apparent. The Full Model, which encompasses all proposed modules (Baseline + LP + V + W), registers the highest performance metrics across all evaluated criteria. These results validate the comprehensive advantage of integrating Line Segment Structural Integration Message Passing, Geometry-Guided Visual Graph, and Feature Weighted Fusion Module.
\end{enumerate}

These findings emphasize the critical role each module plays in the overall performance of the Str-L Pose framework, particularly emphasizing the benefits of combining these components to achieve optimal pose estimation accuracy and reliability.

\begin{table}[ht]
\centering
\renewcommand{\arraystretch}{1.2}
\setlength{\tabcolsep}{8pt}{
\caption{Comparison of inference time with several state-of-the-art methods.}}
\label{table5}
\begin{tabular}{|c|c|c|ll}
\cline{1-3}
Method       & Resolution & Total Time &  &  \\ \cline{1-3}
ORB-SLAM2 \cite{r6}    & 376 × 1241 & 60ms       &  &  \\
SC-SfMLearner \cite{r46} & 256 × 832  & 19.4ms     &  &  \\
Zhou et.al. \cite{r27}   & 192 × 640  & 21.7ms     &  &  \\ \cline{1-3}
% Ours          & 376 × 1241 & 14.3ms     &  &  \\
% Ours          & 256 × 832   & 14.2ms     &  &  \\
Ours          & 192 × 640  & 13.1ms     &  &  \\ \cline{1-3}
\end{tabular}

\end{table}
\subsection{Inference Time Comparison}

To evaluate the performance of our method in terms of inference time, we selected the classic ORB-SLAM2 \cite{r6} , SC-SfMLearner \cite{r46} and the method of Zhou et al. \cite{r27} as comparison methods, and conducted the inference time test on the KITTI Odometry data set. Our methods, ORB-SLAM2 \cite{r6} , SC-SfMLearner \cite{r46} are all tested in the NVIDIA RTX 2080 GPU hardware environment. Zhou et al.'s \cite{r27} method is not open source, so we directly use the results of their paper, and their method is tested on the TITAN RTX GPU. Each method was run 100 times to ensure the stability of the results, and the average inference time (unit: ms) was calculated.

The experimental results are shown in Table \ref{table5}.ORB-SLAM2 \cite{r6} has a relatively long inference time due to its feature point-based SLAM framework, which includes multiple complex steps such as feature extraction, feature matching, and local map optimization. Although ORB-SLAM2 \cite{r6} can provide high-precision pose estimation, its complex calculation process significantly reduces inference speed.

As a self-supervised method based on deep learning, SC-SfMLearner \cite{r46} combines depth estimation and motion estimation during the inference process. Although the estimated accuracy is good, its inference time is relatively high due to the complex network structure, especially the need to process depth information.

The method of Zhou et al. \cite{r27} needs to process point cloud data, and the inference time is slower than our method. In comparison, our method exhibits better inference time. It is attributed to the fact that we do not use complex depth or 3D point cloud information as constraints for pose estimation but use 2D information in the image, and the bottleneck of inference time is mainly in point and line feature detection and matching.
\section{Conclusion}
\label{section5}
In conclusion, Str-L Pose demonstrates the ability to integrate structured geometric features with visual features using a Dual-Graph framework for relative pose estimation. This method improves the accuracy of various environments, especially those with structured environments, and highlights the potential of 2D information. The robust performance of our approach, validated against the DeMoN \cite{r10} and KITTI Odometry \cite{r24} datasets, underscores its superiority over contemporary methods. The fusion of visual features with geometric features, enhanced by our Feature Weighted Fusion Module, heralds a paradigm shift in pose estimation strategies, especially in challenging visual environments. 

\section*{Acknowledgements}
This work was supported by  National Natural Science
Foundation of China (No.62172032)

% \section{Code, Data, and Materials Availability}
% The availability of code, data and materials will be available at \url{https://github.com/oldmanR/Str-L-Pose/}.
%%%%% References %%%%%

\bibliography{report}   % bibliography data in report.bib

\begin{thebibliography}{10}

\bibitem{r62}
M.~Xu, Y.~Wang, B.~Xu, {\em et~al.}, ``A critical analysis of image-based camera pose estimation techniques,'' {\em Neurocomputing} {\bf 570}, 127125  (2024).

\bibitem{r1}
J.~Engel, V.~Koltun, and D.~Cremers, ``Direct sparse odometry,'' {\em IEEE transactions on pattern analysis and machine intelligence} {\bf 40}(3), 611--625  (2017).

\bibitem{r2}
C.~Zhao, Y.~Tang, Q.~Sun, {\em et~al.}, ``Deep direct visual odometry,'' {\em IEEE transactions on intelligent transportation systems} {\bf 23}(7), 7733--7742  (2021).

\bibitem{r3}
J.~Engel, T.~Sch{\"o}ps, and D.~Cremers, ``Lsd-slam: Large-scale direct monocular slam,'' in {\em European conference on computer vision},  834--849, Springer  (2014).

\bibitem{r4}
C.~Forster, Z.~Zhang, M.~Gassner, {\em et~al.}, ``Svo: Semidirect visual odometry for monocular and multicamera systems,'' {\em IEEE Transactions on Robotics} {\bf 33}(2), 249--265  (2016).

\bibitem{r5}
D.~Nist{\'e}r, ``An efficient solution to the five-point relative pose problem,'' {\em IEEE transactions on pattern analysis and machine intelligence} {\bf 26}(6), 756--770  (2004).

\bibitem{r6}
R.~Mur-Artal and J.~D. Tard{\'o}s, ``Orb-slam2: An open-source slam system for monocular, stereo, and rgb-d cameras,'' {\em IEEE transactions on robotics} {\bf 33}(5), 1255--1262  (2017).

\bibitem{r7}
B.~Triggs, P.~F. McLauchlan, R.~I. Hartley, {\em et~al.}, ``Bundle adjustment—a modern synthesis,'' in {\em Vision Algorithms: Theory and Practice: International Workshop on Vision Algorithms Corfu, Greece, September 21--22, 1999 Proceedings},  298--372, Springer  (2000).

\bibitem{r8}
A.~Kendall, M.~Grimes, and R.~Cipolla, ``Posenet: A convolutional network for real-time 6-dof camera relocalization,'' in {\em Proceedings of the IEEE international conference on computer vision},  2938--2946  (2015).

\bibitem{r9}
S.~Sinha, J.~Y. Zhang, A.~Tagliasacchi, {\em et~al.}, ``Sparsepose: Sparse-view camera pose regression and refinement,'' in {\em Proceedings of the IEEE/CVF Conference on Computer Vision and Pattern Recognition},  21349--21359  (2023).

\bibitem{r10}
B.~Ummenhofer, H.~Zhou, J.~Uhrig, {\em et~al.}, ``Demon: Depth and motion network for learning monocular stereo,'' in {\em Proceedings of the IEEE conference on computer vision and pattern recognition},  5038--5047  (2017).

\bibitem{r11}
X.~Wei, Y.~Zhang, Z.~Li, {\em et~al.}, ``Deepsfm: Structure from motion via deep bundle adjustment,'' in {\em Computer Vision--ECCV 2020: 16th European Conference, Glasgow, UK, August 23--28, 2020, Proceedings, Part I 16},  230--247, Springer  (2020).

\bibitem{r12}
R.~Clark, M.~Bloesch, J.~Czarnowski, {\em et~al.}, ``Learning to solve nonlinear least squares for monocular stereo,'' in {\em Proceedings of the European Conference on Computer Vision (ECCV)},  284--299  (2018).

\bibitem{r13}
C.~Tang and P.~Tan, ``Ba-net: Dense bundle adjustment network,'' {\em arXiv preprint arXiv:1806.04807}   (2018).

\bibitem{r14}
B.~Zhuang and M.~Chandraker, ``Fusing the old with the new: Learning relative camera pose with geometry-guided uncertainty,'' in {\em Proceedings of the IEEE/CVF Conference on Computer Vision and Pattern Recognition},  32--42  (2021).

\bibitem{r15}
F.~Xue, X.~Wang, S.~Li, {\em et~al.}, ``Beyond tracking: Selecting memory and refining poses for deep visual odometry,'' in {\em Proceedings of the IEEE/CVF conference on computer vision and pattern recognition},  8575--8583  (2019).

\bibitem{r16}
S.~Wang, R.~Clark, H.~Wen, {\em et~al.}, ``Deepvo: Towards end-to-end visual odometry with deep recurrent convolutional neural networks,'' in {\em 2017 IEEE international conference on robotics and automation (ICRA)},  2043--2050, IEEE  (2017).

\bibitem{r17}
W.~Wang, Y.~Hu, and S.~Scherer, ``Tartanvo: A generalizable learning-based vo,'' in {\em Conference on Robot Learning},  1761--1772, PMLR  (2021).

\bibitem{r63}
R.~Zhu, M.~Yang, W.~Liu, {\em et~al.}, ``Deepavo: Efficient pose refining with feature distilling for deep visual odometry,'' {\em Neurocomputing} {\bf 467}, 22--35  (2022).

\bibitem{r18}
J.~Oliensis, ``A critique of structure-from-motion algorithms,'' {\em Computer Vision and Image Understanding} {\bf 80}(2), 172--214  (2000).

\bibitem{r19}
F.~Nie, W.~Zhang, F.~Li, {\em et~al.}, ``Mismatch removal of visual odometry using klt danger-points tracking and suppression,'' in {\em 2019 IEEE International Conference on Advanced Robotics and its Social Impacts (ARSO)},  330--334, IEEE  (2019).

\bibitem{r20}
A.~Bangunharcana, A.~Magd, and K.-S. Kim, ``Dualrefine: Self-supervised depth and pose estimation through iterative epipolar sampling and refinement toward equilibrium,'' in {\em Proceedings of the IEEE/CVF Conference on Computer Vision and Pattern Recognition},  726--738  (2023).

\bibitem{r21}
Z.~Yin and J.~Shi, ``Geonet: Unsupervised learning of dense depth, optical flow and camera pose,'' in {\em Proceedings of the IEEE conference on computer vision and pattern recognition},  1983--1992  (2018).

\bibitem{r22}
T.~Zhou, M.~Brown, N.~Snavely, {\em et~al.}, ``Unsupervised learning of depth and ego-motion from video,'' in {\em Proceedings of the IEEE conference on computer vision and pattern recognition},  1851--1858  (2017).

\bibitem{r23}
R.~Pautrat, I.~Su{\'a}rez, Y.~Yu, {\em et~al.}, ``Gluestick: Robust image matching by sticking points and lines together,'' in {\em Proceedings of the IEEE/CVF International Conference on Computer Vision},  9706--9716  (2023).

\bibitem{r25}
J.~Bian, Z.~Li, N.~Wang, {\em et~al.}, ``Unsupervised scale-consistent depth and ego-motion learning from monocular video,'' {\em Advances in neural information processing systems} {\bf 32}  (2019).

\bibitem{r26}
R.~Gao, X.~Xiao, W.~Xing, {\em et~al.}, ``Unsupervised learning of monocular depth and ego-motion in outdoor/indoor environments,'' {\em IEEE Internet of Things Journal} {\bf 9}(17), 16247--16258  (2022).

\bibitem{r27}
B.~Zhou, J.~Xie, Z.~Jin, {\em et~al.}, ``Geometry-aware network for unsupervised learning of monocular camera’s ego-motion,'' {\em IEEE Transactions on Intelligent Transportation Systems}   (2023).

\bibitem{r66}
X.~Wei, Y.~Zhang, X.~Ren, {\em et~al.}, ``Deepsfm: Robust deep iterative refinement for structure from motion,'' {\em IEEE Transactions on Pattern Analysis and Machine Intelligence}   (2023).

\bibitem{r28}
F.~Xue, X.~Wu, S.~Cai, {\em et~al.}, ``Learning multi-view camera relocalization with graph neural networks,'' in {\em 2020 IEEE/CVF Conference on Computer Vision and Pattern Recognition (CVPR)},  11372--11381, IEEE  (2020).

\bibitem{r29}
M.~Taiana, M.~Toso, S.~James, {\em et~al.}, ``Posernet: Refining relative camera poses exploiting object detections,'' in {\em European Conference on Computer Vision},  247--263, Springer  (2022).

\bibitem{r30}
E.~Panagiotaki, D.~De~Martini, G.~Pramatarov, {\em et~al.}, ``Sem-gat: Explainable semantic pose estimation using learned graph attention,'' in {\em 2023 21st International Conference on Advanced Robotics (ICAR)},  367--374, IEEE  (2023).

\bibitem{r57}
J.~Wang, C.~Rupprecht, and D.~Novotny, ``Posediffusion: Solving pose estimation via diffusion-aided bundle adjustment,'' in {\em Proceedings of the IEEE/CVF International Conference on Computer Vision},  9773--9783  (2023).

\bibitem{r58}
K.~Chen, N.~Snavely, and A.~Makadia, ``Wide-baseline relative camera pose estimation with directional learning,'' in {\em Proceedings of the IEEE/CVF Conference on Computer Vision and Pattern Recognition},  3258--3268  (2021).

\bibitem{r59}
Y.~Cho, S.~Eum, J.~Im, {\em et~al.}, ``Deep photo-geometric loss for relative camera pose estimation,'' {\em IEEE Access}   (2023).

\bibitem{r60}
K.~Leng, C.~Yang, W.~Sui, {\em et~al.}, ``Sitpose: A siamese convolutional transformer for relative camera pose estimation,'' in {\em 2023 IEEE International Conference on Multimedia and Expo (ICME)},  1871--1876, IEEE  (2023).

\bibitem{r61}
R.~Song, R.~Zhu, Z.~Xiao, {\em et~al.}, ``Contextavo: Local context guided and refining poses for deep visual odometry,'' {\em Neurocomputing} {\bf 533}, 86--103  (2023).

\bibitem{r65}
F.~Fu, J.~Yang, J.~Zhang, {\em et~al.}, ``Eliminating short-term dynamic elements for robust visual simultaneous localization and mapping using a coarse-to-fine strategy,'' {\em Journal of Electronic Imaging} {\bf 31}(5), 053018--053018  (2022).

\bibitem{r33}
A.~Schmied, T.~Fischer, M.~Danelljan, {\em et~al.}, ``R3d3: Dense 3d reconstruction of dynamic scenes from multiple cameras,'' in {\em Proceedings of the IEEE/CVF International Conference on Computer Vision},  3216--3226  (2023).

\bibitem{r64}
X.~Meng, C.~Fan, Y.~Ming, {\em et~al.}, ``Un-vdnet: unsupervised network for visual odometry and depth estimation,'' {\em Journal of Electronic Imaging} {\bf 28}(6), 063015--063015  (2019).

\bibitem{r34}
S.~Shen, Y.~Cai, W.~Wang, {\em et~al.}, ``Dytanvo: Joint refinement of visual odometry and motion segmentation in dynamic environments,'' in {\em 2023 IEEE International Conference on Robotics and Automation (ICRA)},  4048--4055, IEEE  (2023).

\bibitem{r32}
T.~Xie, K.~Dai, S.~Lu, {\em et~al.}, ``Ofvl-ms: Once for visual localization across multiple indoor scenes,'' in {\em Proceedings of the IEEE/CVF International Conference on Computer Vision},  5516--5526  (2023).

\bibitem{r31}
D.~Moran, H.~Koslowsky, Y.~Kasten, {\em et~al.}, ``Deep permutation equivariant structure from motion,'' in {\em Proceedings of the IEEE/CVF International Conference on Computer Vision},  5976--5986  (2021).

\bibitem{r67}
L.~Lai, Z.~Shangguan, J.~Zhang, {\em et~al.}, ``Xvo: Generalized visual odometry via cross-modal self-training,'' in {\em Proceedings of the IEEE/CVF International Conference on Computer Vision},  10094--10105  (2023).

\bibitem{r68}
Y.~Feng, Z.~Guo, Q.~Chen, {\em et~al.}, ``Scipad: Incorporating spatial clues into unsupervised pose-depth joint learning,'' {\em arXiv preprint arXiv:2407.05283}   (2024).

\bibitem{r35}
R.~Gomez-Ojeda, F.-A. Moreno, D.~Zuniga-No{\"e}l, {\em et~al.}, ``Pl-slam: A stereo slam system through the combination of points and line segments,'' {\em IEEE Transactions on Robotics} {\bf 35}(3), 734--746  (2019).

\bibitem{r37}
C.~Qiao, T.~Bai, Z.~Xiang, {\em et~al.}, ``Superline: A robust line segment feature for visual slam,'' in {\em 2021 IEEE/RSJ International Conference on Intelligent Robots and Systems (IROS)},  5664--5670, IEEE  (2021).

\bibitem{r38}
S.~Gao, J.~Wan, Y.~Ping, {\em et~al.}, ``Pose refinement with joint optimization of visual points and lines,'' in {\em 2022 IEEE/RSJ International Conference on Intelligent Robots and Systems (IROS)},  2888--2894, IEEE  (2022).

\bibitem{r36}
H.~Li, J.~Zhao, J.-C. Bazin, {\em et~al.}, ``Hong kong world: Leveraging structural regularity for line-based slam,'' {\em IEEE Transactions on Pattern Analysis and Machine Intelligence}   (2023).

\bibitem{r39}
F.~Liu, M.~Huang, H.~Ge, {\em et~al.}, ``Unsupervised monocular depth estimation for monocular visual slam systems,'' {\em IEEE Transactions on Instrumentation and Measurement}   (2023).

\bibitem{r56}
A.~Kendall and Y.~Gal, ``What uncertainties do we need in bayesian deep learning for computer vision?,'' {\em Advances in neural information processing systems} {\bf 30}  (2017).

\bibitem{r53}
S.~Fuhrmann, F.~Langguth, and M.~Goesele, ``Mve-a multi-view reconstruction environment.,'' {\em GCH} {\bf 3}, 4  (2014).

\bibitem{r52}
J.~Xiao, A.~Owens, and A.~Torralba, ``Sun3d: A database of big spaces reconstructed using sfm and object labels,'' in {\em Proceedings of the IEEE international conference on computer vision},  1625--1632  (2013).

\bibitem{r54}
J.~Sturm, N.~Engelhard, F.~Endres, {\em et~al.}, ``A benchmark for the evaluation of rgb-d slam systems,'' in {\em 2012 IEEE/RSJ international conference on intelligent robots and systems},  573--580, IEEE  (2012).

\bibitem{r55}
A.~X. Chang, T.~Funkhouser, L.~Guibas, {\em et~al.}, ``Shapenet: An information-rich 3d model repository,'' {\em arXiv preprint arXiv:1512.03012}   (2015).

\bibitem{r24}
A.~Geiger, P.~Lenz, and R.~Urtasun, ``Are we ready for autonomous driving? the kitti vision benchmark suite,'' in {\em Conference on Computer Vision and Pattern Recognition (CVPR)},   (2012).

\bibitem{r40}
K.~M. Yi, E.~Trulls, Y.~Ono, {\em et~al.}, ``Learning to find good correspondences,'' in {\em Proceedings of the IEEE conference on computer vision and pattern recognition},  2666--2674  (2018).

\bibitem{r41}
C.~Zhao, Z.~Cao, C.~Li, {\em et~al.}, ``Nm-net: Mining reliable neighbors for robust feature correspondences,'' in {\em Proceedings of the IEEE/CVF conference on computer vision and pattern recognition},  215--224  (2019).

\bibitem{r42}
R.~Li, S.~Wang, Z.~Long, {\em et~al.}, ``Undeepvo: Monocular visual odometry through unsupervised deep learning,'' in {\em 2018 IEEE international conference on robotics and automation (ICRA)},  7286--7291, IEEE  (2018).

\bibitem{r43}
H.~Zhan, R.~Garg, C.~S. Weerasekera, {\em et~al.}, ``Unsupervised learning of monocular depth estimation and visual odometry with deep feature reconstruction,'' in {\em Proceedings of the IEEE conference on computer vision and pattern recognition},  340--349  (2018).

\bibitem{r44}
R.~Wang, S.~M. Pizer, and J.-M. Frahm, ``Recurrent neural network for (un-) supervised learning of monocular video visual odometry and depth,'' in {\em Proceedings of the IEEE/CVF Conference on Computer Vision and Pattern Recognition},  5555--5564  (2019).

\bibitem{r45}
C.~Godard, O.~Mac~Aodha, M.~Firman, {\em et~al.}, ``Digging into self-supervised monocular depth estimation,'' in {\em Proceedings of the IEEE/CVF international conference on computer vision},  3828--3838  (2019).

\bibitem{r46}
J.~Bian, Z.~Li, N.~Wang, {\em et~al.}, ``Unsupervised scale-consistent depth and ego-motion learning from monocular video,'' {\em Advances in neural information processing systems} {\bf 32}  (2019).

\bibitem{r47}
A.~Ranjan, V.~Jampani, L.~Balles, {\em et~al.}, ``Competitive collaboration: Joint unsupervised learning of depth, camera motion, optical flow and motion segmentation,'' in {\em Proceedings of the IEEE/CVF conference on computer vision and pattern recognition},  12240--12249  (2019).

\bibitem{r48}
C.~Shu, K.~Yu, Z.~Duan, {\em et~al.}, ``Feature-metric loss for self-supervised learning of depth and egomotion,'' in {\em European Conference on Computer Vision},  572--588, Springer  (2020).

\bibitem{r49}
W.~Mai and Y.~Watanabe, ``Feature-aided bundle adjustment learning framework for self-supervised monocular visual odometry,'' in {\em 2021 IEEE/RSJ International Conference on Intelligent Robots and Systems (IROS)},  9160--9167, IEEE  (2021).

\bibitem{r50}
S.~Lee, S.~Im, S.~Lin, {\em et~al.}, ``Learning monocular depth in dynamic scenes via instance-aware projection consistency,'' in {\em Proceedings of the AAAI conference on artificial intelligence},   {\bf 35}(3), 1863--1872  (2021).

\bibitem{r51}
Z.~Liang, Q.~Wang, and Y.~Yu, ``Deep unsupervised learning based visual odometry with multi-scale matching and latent feature constraint,'' in {\em 2021 IEEE/RSJ International Conference on Intelligent Robots and Systems (IROS)},  2239--2246, IEEE  (2021).

\bibitem{r69}
H.~Wang, C.~Wang, C.-L. Chen, {\em et~al.}, ``F-loam: Fast lidar odometry and mapping,'' in {\em 2021 IEEE/RSJ International Conference on Intelligent Robots and Systems (IROS)},  4390--4396, IEEE  (2021).

\bibitem{r70}
Y.~Pan, X.~Zhong, L.~Wiesmann, {\em et~al.}, ``Pin-slam: Lidar slam using a point-based implicit neural representation for achieving global map consistency,'' {\em arXiv preprint arXiv:2401.09101}   (2024).

\end{thebibliography}
\bibliographystyle{spiejour}   % makes bibtex use spiejour.bst
\vspace{2ex}\noindent\textbf{Zherong Zhang} received a Bachelor's degree in Internet of Things Engineering from Beijing Jiaotong University in China and is currently pursuing a Master's degree at the Institute of Information Science. His current research interests include camera calibration and camera pose estimation.

\vspace{2ex}\noindent\textbf{Chunyu Lin} received the Ph.D. degree from Beijing Jiaotong University (BJTU), Beijing, China, in 2011. From 2009 to 2010, he was a Visiting Researcher with the ICT Group, Delft University of Technology, The Netherlands. From 2011 to 2012, he was a Postdoctoral Researcher with the Multimedia Laboratory, Gent University, Belgium. He is currently a Professor with BJTU. His research interests include image/video compression, 3D vision, virtual reality video processing, and ADAS.

\vspace{2ex}\noindent\textbf{Shangrong Yang} received the B.S degree from Beijing Jiaotong University (BJTU), Beijing, China, in 2018, where he is currently pursuing the Ph.D. degree in signal and information processing with the Institute of Information Science. His current research interests focuses on low-level vision, including: image generation/restoration/enhancement, video processing, fisheye/panoramic distortion rectification, inpainting and outpainting, object detection and tracking, etc.

\vspace{2ex}\noindent\textbf{Shujuan Huang} received the B.S. degree in the Internet of Things engineering from Beijing Jiaotong University, Beijing, China, where he is currently pursuing the Ph.D. degree with the Institute of Information Science. His current research interests include camera calibration, pose estimation and vehicle localization.

\vspace{2ex}\noindent\textbf{Yao Zhao} received his Ph.D. from Beijing Jiaotong University in 1996. He became a professor in 2001 and serves as director of the Institute of Information Science at BJTU. His research focuses on image coding, digital watermarking \& forensics, cross-media content analysis, and multimedia information processing. He worked as a Postdoc at Delft University from 2001 to 2002 and was named a Chang Jiang Scholar in 2013.
% \vspace{1ex}
% \noindent Biographies and photographs of the other authors are not available.

\listoffigures
\listoftables

\end{spacing}
\end{document}